\documentclass[journal]{IEEEtran}
\usepackage{cite}
\usepackage{graphicx}
\usepackage{booktabs}
\usepackage{multirow}
\usepackage{bm}
\usepackage{amsmath}
\ifCLASSINFOpdf
\else
\fi
\begin{document}
  %
  \title{Spiking Neural-Invariant Kalman Fusion for Accurate Localization Using Low-Cost IMUs}
  %
  %
  %

  
  \author{Yaohua~Liu, Qiao~Xu, Binkai~Ou 
  \thanks{Y. Liu is with Guangdong Institute of Intelligence Science and Technology,
  Hengqin, Zhuhai, Guangdong, China, 519031, e-mail: liuyaohua@gdiist.cn.}
  \thanks{Q. Xu is with East China Normal University, Shanghai, 200062, China}
  \thanks{B. Ou is with Innovation and Research and Development Department, BoardWare Information System Co.Ltd, Macau, 999078, China.}
  }
  \maketitle

  \begin{abstract}    
 Low-cost inertial measurement units (IMUs) are widely utilized in mobile robot localization due to their affordability and ease of integration. However, their complex, nonlinear, and time-varying noise characteristics often lead to significant degradation in localization accuracy when applied directly for dead reckoning. To overcome this limitation, we propose a novel brain-inspired state estimation framework that combines a spiking neural network (SNN) with an invariant extended Kalman filter (InEKF). The SNN is designed to extract motion-related features from long sequences of IMU data affected by substantial random noise and is trained via a surrogate gradient descent algorithm to enable dynamic adaptation of the covariance noise parameter within the InEKF. By fusing the SNN output with raw IMU measurements, the proposed method enhances the robustness and accuracy of pose estimation. Extensive experiments conducted on the KITTI dataset and real-world data collected using a mobile robot equipped with a low-cost IMU demonstrate that the proposed approach outperforms state-of-the-art methods in localization accuracy and exhibits strong robustness to sensor noise, highlighting its potential for real-world mobile robot applications.
  \end{abstract}


  %
  \IEEEpeerreviewmaketitle

  \section{Introduction}
  \IEEEPARstart{A}{ccurate} localization is a fundamental requirement for intelligent vehicles, enabling them to determine their precise position within a given environment. The Global Navigation Satellite System (GNSS) is widely employed for this purpose, offering continuous, centimeter-level positioning. However, GNSS is inherently limited by a low update rate and is highly susceptible to multipath effects and signal loss in dense urban areas or enclosed environments such as tunnels. Vision-based localization systems, which rely on cameras or LiDAR, also face considerable challenges in scenarios with poor lighting or insufficient visual features. In contrast, inertial measurement units (IMUs) provide attitude and position estimates by integrating measurements from three-axis gyroscopes and accelerometers, making them an essential component in mobile robotic systems. When combined with other sensors, IMUs can deliver high-frequency inertial data and maintain pose estimation during temporary failures of external sensors. While high-precision IMUs offer excellent performance, their cost renders them impractical for commercial autonomous robots. As an alternative, low-cost IMUs based on micro-electromechanical systems (MEMS) technology have become increasingly popular due to their compact size and energy efficiency. Nonetheless, these sensors exhibit complex, nonlinear, and time-varying noise characteristics, which significantly degrade localization accuracy when directly used for dead reckoning.

  Bayesian filters, particularly various forms of the Kalman filter, are commonly used to integrate IMU data for improved pose estimation during dead reckoning. This process relies on real-time estimation of a vehicle's position, velocity, and orientation by integrating accelerometer and gyroscope measurements through kinematic models. To mitigate inevitable error accumulation from IMU integration, Kalman filters dynamically fuse predictive and observational data, with uncertainty modeled through a covariance noise matrix. However, specifying this parameter poses two major challenges: it is susceptible to system observability, making manual tuning difficult, and it requires real-time adaptation to accommodate nonlinear motion dynamics, such as amplified lateral slip variance during curved trajectories compared to straight-line motion. To address these issues, AI-enhanced methods have been proposed to learn and adapt the noise parameters for low-cost IMUs automatically \cite{brossard2020ai, guo2023model, zhou2022imu}. Nevertheless, most existing approaches rely on convolutional or recurrent neural networks, which are often inadequate in extracting robust motion features from long IMU sequences corrupted by substantial random noise.

 Spiking neural networks (SNNs), regarded as the third generation of artificial neural networks, are inspired by the spatiotemporal dynamics of biological neurons and offer a distinctive framework for processing temporal information. Recent studies have demonstrated that SNNs can achieve performance comparable to traditional deep learning models in various time-series tasks \cite{yao2023attention, fang2020multivariate, lv2024efficient}. In the context of dead reckoning, which inherently involves sequential IMU measurements over time windows for state estimation, SNNs are well-suited to capture temporal patterns and extract spatial features from noisy IMU data. This enables the real-time adaptation of noise parameters, a capability particularly beneficial for low-cost MEMS IMUs that are prone to significant noise and bias \cite{stagsted2020towards}. By leveraging the temporal coding and event-driven characteristics of SNNs, the robustness and accuracy of state estimation in dead reckoning can be significantly improved, especially in challenging environments where conventional methods often fail.

  In this paper, we propose a hybrid state estimation framework that integrates an SNN with an invariant extended Kalman filter (InEKF) to enable accurate dead reckoning using low-cost MEMS IMUs. The SNN is designed to extract spatial features from IMU data and dynamically adjust the noise covariance parameters of the InEKF in real-time, all with minimal computational overhead. The InEKF maintains invariance to certain group transformations, such as rotations and translations, thereby improving robustness under noisy and uncertain conditions. The main contributions of this paper are
  summarized as follows:
  \begin{itemize}
    \item We propose a brain-inspired fusion method that integrates an SNN with an InEKF to achieve accurate dead reckoning for low-cost MEMS IMUs. To the best of our knowledge, this is the first work that combines SNNs with InEKF for IMU dead reckoning. In comparison to traditional deep learning methods, the proposed method is more efficient in
      processing temporal information and can extract spatial features from low-cost
      MEMS IMU data with minimal computational cost.

    \item A hybrid state estimation strategy is developed in which the SNN not only performs feature extraction but also enables adaptive noise modeling, while the InEKF provides consistent and transformation-invariant state estimation

    \item Extensive experiments on the KITTI odometry benchmark and real field tests demonstrate that the proposed method significantly improves localization accuracy and robustness, particularly in degraded sensor scenarios such as IMU drift or GPS signal loss, validating the effectiveness of the approach for practical low-cost inertial navigation applications.
  \end{itemize}

  The remainder of this paper is organized as follows. Section II reviews related work on IMU-based dead reckoning and recent AI-enhanced IMU methods. Section III introduces the IMU error model, the underlying kinematic model, and the theoretical foundations of the InEKF and SNNs. Section IV details the proposed hybrid dead reckoning framework, including SNN-based feature extraction and noise parameter adaptation, as well as InEKF-based state estimation. Section V presents experimental results and analysis on the KITTI odometry dataset and real-world field tests, validating the performance and robustness of the proposed method. Finally, Section VI concludes the paper and outlines directions for future research.

  \section{Related Works}
  Errors in low-cost MEMS IMUs are typically substantial and arise from various sources, including bias instability, scale factor variations, and sensor misalignment. To enhance the accuracy of such IMUs, calibration is generally necessary. Classical calibration methods in inertial navigation are broadly categorized into discrete and system-level approaches \cite{huang2022mems}. Discrete calibration methods, such as the Allan variance technique \cite{allan2005statistics}, estimate IMU errors by analyzing the statistical characteristics of sensor outputs. The Kalibr library \cite{furgale2012continuous} is a widely adopted toolkit for offline calibration, capable of estimating IMU intrinsic parameters and camera-IMU extrinsic parameters. Lu et al. \cite{lu2022all} proposed a comprehensive calibration method for six-axis skewed IMUs using a two-step procedure with complexity comparable to conventional techniques. In another study \cite{lin2024pedestrian}, a multi-IMU array system with weighted fusion significantly reduced measurement noise and improved accuracy. However, discrete calibration methods typically require specialized equipment such as precision turntables and involve high operational costs, making them impractical for the mass deployment of low-cost MEMS IMUs.

 System-level calibration approaches primarily leverage Kalman filter-based techniques \cite{yang2023online}, which estimate IMU errors by fusing IMU measurements with data from complementary sensors such as GPS or LiDAR. For example, Jung et al. \cite{jung2020observability} proposed a self-calibrating visual-inertial odometry (VIO) system that employs an extended Kalman filter (EKF) to estimate IMU scale factor and misalignment errors. To improve robustness during external sensor outages, the Zero-Velocity Update (ZUPT) technique has been introduced to enhance EKF performance by incorporating instantaneous velocity constraints as pseudo-observations \cite{dissanayake2002aiding}. Building on Lie group theory, Barrau and Bonnabel \cite{barrau2016invariant} developed the InEKF, which incorporates geometric invariance principles to maintain consistent and robust state estimation, particularly for wheeled robots operating under nonholonomic constraints. While system-level calibration methods reduce the reliance on specialized equipment like turntables, they typically assume a simplified IMU error model and struggle to capture its inherent nonlinearities. Consequently, achieving high-precision localization using such model-based methods often requires either highly accurate system modeling or meticulous parameter tuning—both of which are challenging and costly in the context of low-cost MEMS IMUs.

 In recent years, AI-enhanced IMU methods have been proposed to automatically learn IMU error characteristics and adapt noise parameters dynamically for low-cost MEMS IMUs. These approaches leverage deep learning models to extract features from raw IMU data and adjust the noise covariance of Kalman filters in real-time. For instance, Brossard et al. \cite{brossard2020ai} introduced a convolutional neural network (CNN)-based framework to adaptively tune filter parameters from IMU inputs, while Zhou et al. \cite{zhou2022imu} employed a recurrent neural network (RNN) to model temporal dependencies in IMU errors. Guo et al. \cite{guo2023model} proposed a hybrid approach that combines model-based constraints with learning-based adaptation to enhance dead reckoning accuracy in wheeled robots. Although these methods demonstrate promising results in improving localization performance using low-cost IMUs, their reliance on CNNs and RNNs limits their ability to capture long-term dependencies and extract motion-relevant features from IMU sequences heavily corrupted by random noise. To address this, Guyard et al. \cite{guyard2025transformer} proposed a Transformer-based bidirectional encoder for IMU-GPS fusion during GPS outages, showing improved modeling of long-term temporal patterns. However, such deep learning methods typically require large-scale, high-quality datasets, which constrain their generalization in low-data regimes. In contrast, SNNs offer greater efficiency and robustness in temporal feature extraction and can process low-cost MEMS IMU data with minimal computational cost \cite{lv2024efficient}. Compared to the aforementioned work, our proposed hybrid state estimation method that integrates a spike neural network with an invariant extended Kalman filter can achieve more accurate dead reckoning with a single low-cost MEMS IMU.

  \section{Low-Cost MEMS IMU and Dead Reckoning Modeling}
 \subsection{Low-Cost MEMS IMU Error Model}
 The IMU is a critical sensor in mobile robot navigation, providing measurements of angular velocity and linear acceleration. However, IMU data are affected by various error sources, including bias, scale factor distortion, sensor misalignment, and stochastic noise. These inaccuracies can significantly degrade the reliability and accuracy of state estimation in mobile robotic systems. To account for these effects, we model the IMU measurement errors by considering the following contributing factors:
  \begin{itemize}
    \item \textbf{Bias}: Bias refers to a constant or slow time-varying error that systematically affects IMU measurements. It can arise from various sources, including temperature fluctuations, mechanical stress, and imperfections introduced during manufacturing. In practice, IMU bias is commonly modeled as a random walk process, wherein the bias at time t is assumed to be equal to the bias at time t-1 perturbed by a zero-mean Gaussian noise term. This formulation captures the gradual and stochastic nature of bias drift over time.

    \item \textbf{Scale factor}: The scale factor represents a multiplicative error that affects the amplitude of IMU measurements. Similar to bias, it can result from temperature variations, mechanical stress, or manufacturing imperfections. In most cases, the scale factor is modeled as a constant or slowly time-varying parameter, reflecting its relatively stable yet imperfect calibration over time. This error leads to proportional distortion in the measured angular velocity and linear acceleration, thereby impacting the accuracy of state estimation if uncorrected.

    \item \textbf{Misalignment}: Misalignment refers to angular deviations between the actual sensor axes and the ideal reference axes, resulting in cross-axis sensitivity and distortion in IMU measurements. This error is typically introduced by mechanical stress, structural deformation, or manufacturing imperfections. Misalignment is commonly modeled as a constant or slowly time-varying parameter, as its variation tends to be gradual over time. If not properly compensated, it can lead to significant degradation in the accuracy of attitude and position estimation.

    \item \textbf{Noise}: Random noise is an inherent stochastic error that affects IMU measurements and arises from various sources, including thermal fluctuations, quantization effects, and electromagnetic interference. This type of noise is typically modeled as zero-mean Gaussian white noise, characterized by a predefined covariance matrix that encapsulates the uncertainty of the sensor. Accurate modeling of this noise is essential for reliable state estimation, as it directly influences the performance of filtering and sensor fusion algorithms.
  \end{itemize}

  Based on the above factors, the IMU measurements and the true values can be
  described as, see \cite{rehder2016extending}, \cite{rohac2015calibration},

  \begin{equation}
    {\bm{\tilde \omega }}_{t}^{{IMU}}={\bf{C}}_{_{1}}^{T}{{\bm{\omega }}_{t}}+{\bf{b}}
    _{t}^{\omega }+{\bm{\eta }}_{t}^{\omega }\label{eq:omega_measurement}
  \end{equation}

  \begin{equation}
    {\bf{\tilde a}}_{t}^{{IMU}}={\bf{C}}_{_{2}}^{T}{\bf{{a}}_{t}}+{\bf{b}}_{t}^{a}
    +{\bm{\eta }}_{t}^{a}\label{eq:a_measurement}
  \end{equation}
  where ${\bm{\tilde \omega }}_{t}^{{IMU}}$ and ${\bf{\tilde a}}_{t}^{{IMU}}$ are
  the angular velocity and linear acceleration measurements of the IMU at time $t$,
  respectively; ${{\bm{\omega }}_{t}}$ and ${{\bf{a}}_{t}}$ denote the true angular
  velocity and linear acceleration, respectively; ${\bf{b}}_{t}^{\omega }$ and ${\bf{b}}
  _{t}^{a}$ represent the gyroscope and accelerometer biases, respectively; ${\bm{\eta }}
  _{t}^{\omega}$ and ${\bm{\eta }}_{t}^{a}$ are the gyroscope and accelerometer noise,
  and they can be regarded as zero-mean Gaussian noise; ${\bf{C}}_{_{1}}^{T}$ and
  ${\bf{C}}_{_{2}}^{T}$ are the calibration matrix of gyroscope and accelerometer,
  respectively. Therefore, the calibration matrix {\bf{C}} can be further
  represented as,
  \begin{equation}
    {\bf{C}}= \left[{\begin{array}{*{20}{c}}{{\bf{C}}_{1}^{T}}\\ {{\bf{C}}_{2}^{T}}\end{array}}
    \right] = \left[{\begin{array}{*{20}{c}}{{{\bf{S}}_{\omega}}{{\bf{M}}_{\bf{\omega }}}}&{\bf{A}}\\ {{{\bf{0}}_{{3 \times 3}}}}&{{{\bf{S}}_{a}}{{\bf{M}}_{a}}}\end{array}}
    \right]
  \end{equation}
  where ${{{\bf{S}}_{\omega}}}$ and ${{{\bf{S}}_{a}}}$ are the scale factor of gyroscope
  and accelerometer, respectively; ${{{\bf{M}}_{\omega}}}$ and ${{{\bf{M}}_{a}}}$
  denote the axis misalignments, and ${\bf{A}}$ is the coefficient matrix
  representing the cross-coupling effects between the gyroscope and
  accelerometer, a.k.a. g-sensitivity.

  \subsection{Kinematic Model of the Mobile Robot}
  The kinematic model of a mobile robot characterizes the relationship between its motion and state variables, including attitude, velocity, and position. In this study, we focus on a wheeled mobile robot subject to nonholonomic constraints, which restrict its motion to the forward or backward direction along its instantaneous heading. Under this assumption, the attitude increment of the robot can be described as follows \cite{huang2022mems},
  \begin{equation}
    {\bf{R}}_{t}^{{IMU}}={\bf{R}}_{{t - 1}}^{{IMU}}\exp ({{\bm{\theta }}_{{t - 1}}}
    ) \label{eq:angular}
  \end{equation}
  where ${\bf{R}}_{t}^{{IMU}}$ is the rotation matrix of the mobile robot at
  time $t$, i.e., that maps the IMU frame to the world frame;
  ${\bf{R}}_{{t - 1}}^{{IMU}}$ is the rotation matrix of the mobile robot at
  time $t-1$; and ${\bm{\theta }}_{{t - 1}}$ is the angular velocity vector of the
  mobile robot at time $t-1$, which can be obtained from the gyroscope
  measurements as follows,
  \begin{equation}
    {{\bm{\theta }}_{{t - 1}}}={\bm{\tilde \omega }}_{{t - 1}}^{{IMU}}dt\label{eq:gyro_to_angular}
  \end{equation}
  The $\exp ({{\bm{\theta }}_{{t - 1}}})$ can be further expressed as,
  \begin{equation}
    \exp ({{\bm{\theta }}_{{t - 1}}}) ={\bf{I}}+ \frac{{\sin {\theta _{t - 1}}}}{{{\theta _{t - 1}}}}
    [{\theta _{t - 1}}\times ] + \frac{{1 - \cos {\theta _{t - 1}}}}{{\theta _{t - 1}^2}}
    {[{\theta _{t - 1}} \times ]^2}\label{eq:exp_theta}
  \end{equation}
  where ${\bf{I}}$ is the identity matrix, and $[{\theta _{t - 1}}\times ]$ is the
  skew-symmetric matrix of the angular vector ${\bf{\theta }}_{{t - 1}}$. The
  velocity increment of the mobile robot can be described by the following,
  \begin{equation}
    {{\bf{v}}_{t}}={{\bf{v}}_{{t - 1}}}+ ({{\bf{R}}_{{t - 1}}}{{{\bf{\tilde a}}}_{t - 1}}
    -{\bf{g}})dt\label{eq:velocity}
  \end{equation}
  where ${\bf{v}}_{\bf{t}}$ is the velocity vector of the mobile robot at time $t$,
  and $g$ is gravity.

  The position is updated by integrating the velocity over time,
  \begin{equation}
    {{\bf{p}}_{t}}={{\bf{p}}_{{t - 1}}}+{{\bf{v}}_{{t - 1}}}dt\label{eq:position}
  \end{equation}
  where ${\bf{p}}_{t}$ is the position vector of the mobile robot at time $t$.

  The errors described above propagate through the motion equations Eq.\ref{eq:angular}-Eq.\ref{eq:position}, ultimately affecting the estimated rotation, velocity, and position of the mobile robot, with their impact accumulating over time. Consequently, effective denoising of IMU measurements and accurate state estimation are essential for reliable dead reckoning. To address this, we propose a method that integrates the IMU error model with a spiking Transformer architecture, which enables dynamic adjustment of model parameters and facilitates the extraction of cleaner, denoised IMU signals for improved state estimation.

  \subsection{Invariant Extended Kalman Filter}
 The InEKF is a variant of the classical EKF that leverages geometric invariance principles to enhance the robustness of state estimation in systems subject to nonholonomic constraints \cite{barrau2016invariant}. Unlike the standard EKF, the InEKF is formulated on Lie groups, ensuring consistency and invariance under transformations such as rotations and translations. This property is particularly beneficial for mobile robotic systems. The system state in the InEKF framework can be formulated as follows,
  \begin{equation}
    {\bf{x}}_{t}={f}({\bf{x}}_{{t - 1}},{\bf{u}}_{t - 1}) +{\bf{w}}_{t-1}\label{eq:state_transition}
  \end{equation}
  where ${\bf{x}}_{t}$ is the state vector of the system at time $t$, ${\bf{f}}$
  is the state transition function that describes the dynamics of the system, ${\bf{u}}
  _{t - 1}$ is the control input at time $t-1$, and ${\bf{w}}_{t-1}$ is the
  process noise, which is assumed to be Gaussian with zero mean and a certain covariance
  matrix ${\bf{Q}}_{t-1}$.

  The measurement model of the InEKF can be described as follows,
  \begin{equation}
    {\bf{z}}_{t}={h}({\bf{x}}_{t}) +{\bf{v}}_{t}\label{eq:measurement_model}
  \end{equation}
  where ${\bf{z}}_{\bf{t}}$ is the measurement vector at time $t$, ${\bf{h}}$ is
  the measurement function that maps the state vector to the measurement space,
  and ${\bf{v}}_{t}$ is the measurement noise, which is also assumed to be Gaussian
  with zero mean and a certain covariance matrix ${\bf{R}}_{t}$.

  The InEKF initializes with a Gaussian prior distribution over the state vector,
  denoted as $\mathcal{N}({{{\bf{\hat x}}}_0},{{\bf{P}}_0})$, where ${{{\bf{\hat x}}}_0}$
  is the initial state estimate and ${{\bf{P}}_0}$ is the initial covariance matrix.
  The InEKF then iteratively updates the state estimate and covariance matrix
  based on the incoming measurements and the system dynamics. The process of InEKF
  can be divided into two main steps: prediction and correction. The prediction step
  of the InEKF can be described as follows:
  \begin{equation}
    {{\bf{P}}_t}={{\bf{F}}_{t - 1}}{{\bf{P}}_{t - 1}}{\bf{F}}_{t - 1}^{T}+{{\bf{G}}_{t - 1}}
    {{\bf{Q}}_{t - 1}}{\bf{G}}_{t - 1}^{T}\label{eq:prediction_step}
  \end{equation}
  where ${{\bf{F}}_{t - 1}}$ and ${{\bf{G}}_{t - 1}}$ is the Jacobian matrix of
  the state transition function ${f}$ with respect to the state vector and the
  process noise. During the measurement update phase, the filter incorporates
  pseudo-observations through the Kalman gain mechanism.

  In practical implementations of InEKF, precise specification of the system dynamics
  $f$ and observation model $h$, along with their associated process noise
  covariance ${{\bf{Q}}_t}$ and measurement noise covariance ${{\bf{N}}_t}$, is essential
  for optimal performance. Conventionally, these noise parameters require manual
  tuning, which often presents significant challenges in achieving proper filter
  convergence. To address this limitation, our proposed framework employs a
  neural network to autonomously estimate the complete set of noise parameters
  $\{{{\bf{Q}}_t},{{\bf{N}}_t}\}$ in real-time, thereby eliminating the need for
  manual calibration while maintaining rigorous estimation-theoretic guarantees.

  \subsection{Spiking Neurons and Surrogate Gradient}
\begin{figure}[!ht]
    \centering
    \includegraphics[width=.9\linewidth]{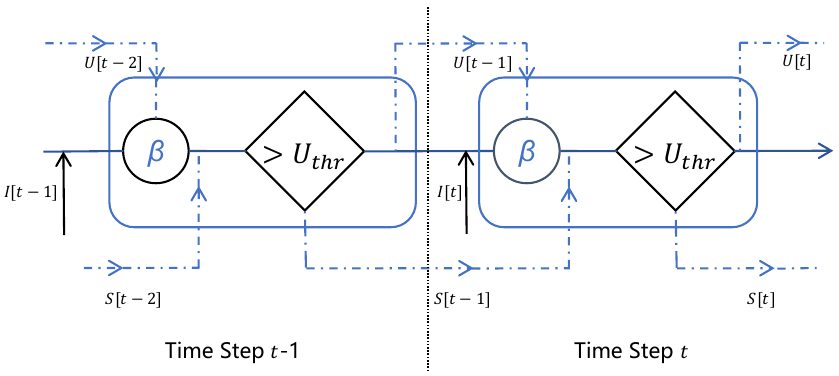}
    \caption{The structure illustration of the LIF neuron.}
    \label{fig:LIF}
  \end{figure}

  Unlike traditional deep neural networks (DNNs), the fundamental computational unit in the SNN used in our proposed method is the leaky integrate-and-fire (LIF) neuron \cite{maass1997networks}, which offers a more biologically plausible representation of neuronal dynamics. As illustrated in Fig. \ref{fig:LIF}, an LIF neuron integrates incoming spikes over time, and emits an output spike when its membrane potential surpasses a predefined threshold. The evolution of the membrane potential in the LIF model can be mathematically described as follows,
  \begin{equation}
    \begin{array}{l}
      U(t)=H(t-\Delta t)+I(t), \quad I(t)=f(\mathbf{x} ; \theta),                                                                         \\
      H(t)=V_{\text{reset }} S(t)+(1-S(t)) \beta U(t),                                                                                    \\
      S(t)=\left\{\begin{array}{ll}1, & \text{ if } U(t) \geq U_{\text{thr }}, \\ 0, & \text{ if } U(t)<U_{\text{thr }}\end{array}\right.
    \end{array}
    \label{eq:LIF}
  \end{equation}

  In the context of the LIF neuron model, the spatial input $I(t)$ at time step $t$
  is computed by applying a learnable function $f$ to the input $\bf{x}$, parameterized by $\theta$. The temporal output $H(t)$ is governed by a discretization constant $\Delta t$, which controls the resolution of the LIF dynamics. The spike response $S(t)$ is defined using a Heaviside step function, triggered when the membrane potential $U(t)$ exceeds a predefined threshold $ U_{\text{thr }}$. Upon firing, the neuron emits a spike and the membrane potential is reset to $V_{reset}$. If the threshold is not reached, no spike is generated, and the membrane potential decays toward $H(t)$ at a rate determined by the decay rate $\beta$. This formulation enables the neuron to capture temporal dependencies and sparse activation patterns, which are crucial for efficient spatiotemporal feature extraction.

  Through a spiking neuron layer $SN( \cdot )$, the spike trains $\bf{S}$ are generated by iterating over ${T'}$ discrete time steps, where each of the $N$ input currents $\bf{I}$ is processed by a corresponding LIF neuron. Formally, this can be expressed as,
  \begin{equation}
    {\bf{S}}= SN({\bf{I}})\label{eq:spike_layer}
  \end{equation}
  where the arctangent-like surrogate gradients \cite{fang2023spikingjelly} are used
  to approximate the gradients of the spiking neurons during the backpropagation
  process. The surrogate gradient is defined as follows,
  \begin{equation}
    S(t) \approx \frac{1}{\pi }\arctan (\frac{\pi }{2}\alpha U(t)) + \frac{1}{2}\label{eq:surrogate_gradient}
  \end{equation}
  where $\alpha$ is a hyperparameter that controls the steepness of the surrogate
  gradient. Therefore, the comprehensive model can be trained using an end-to-end
  approach, facilitated by backpropagation through time (BPTT), which enables efficient
  learning and optimization.

   \section{Proposed Method}
  \subsection{The Hybrid State Estimation Strategy for IMU Dead Reckoning}
  The proposed hybrid state estimation framework integrates an SNN with an InEKF to enable accurate dead reckoning using low-cost MEMS IMUs. In general, IMU dead reckoning errors can be categorized into two primary sources: (1) measurement errors arising from sensor noise and bias, and (2) state estimation errors resulting from uncertainties in system dynamics and measurement models. The objective of the proposed method is to mitigate both error types by leveraging the complementary strengths of data-driven SNNs and model-based InEKF. Notably, dead reckoning with low-cost MEMS IMUs involves nonlinear dynamics and is subject to significant stochastic noise, making it difficult to establish precise analytical models. Traditional filter-based methods rely heavily on the accuracy of system modeling and sensor quality; as a result, their performance often degrades when confronted with corrupted IMU data, such as those affected by mechanical vibration or environmental disturbances. Moreover, such methods typically require manual tuning of filter parameters, including noise covariance matrices, which are labor-intensive and lack adaptability in dynamic environments. In contrast, the proposed framework adopts a data-driven paradigm wherein the SNN learns temporal patterns directly from raw IMU data and adaptively adjusts the noise parameters of the InEKF in real-time. Inspired by the brain’s biological mechanisms for processing temporal information, SNNs exhibit strong capabilities in extracting spatial features from noisy inputs while maintaining high robustness and estimation accuracy, outperforming conventional deep learning models in this context.

\begin{figure*}[!ht]
    \centering
    \includegraphics[width=.7\linewidth]{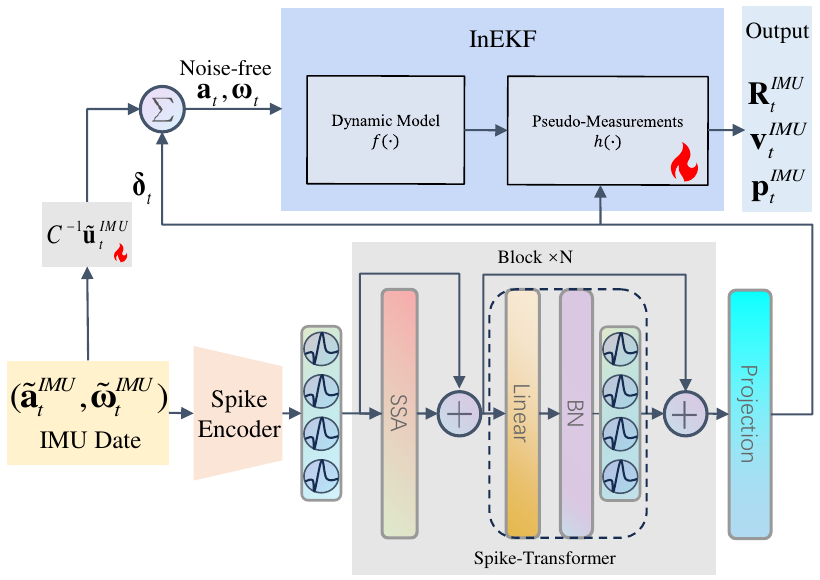}
    \caption{An overview of the proposed hybrid state estimation method for IMU
    dead reckoning. }
    \label{fig:framework}
  \end{figure*}

  As shown in Fig.\ref{fig:framework}, the raw IMU measurements are first calibrated
  using the IMU error model to reduce the bias and noise of the low-cost MEMS IMU.
  Specifically, according to the analysis of the IMU error model in Eq.\ref{eq:omega_measurement}
  and Eq.\ref{eq:a_measurement}, the IMU measurements can be calibrated by
  estimating the calibration matrix ${\bf{C}}$, the biases ${\bf{b}}_{t}^{IMU}$,
  and the noise ${\bm{\eta }}_{t}^{IMU}$. Thus, the IMU measurements after
  calibration can be further expressed as,
  \begin{equation}
    {\bf{u}}_{t}^{IMU}={C^{ - 1}}({\bf{\tilde u}}_{t}^{IMU}-{\bf{b}}_{t}^{IMU}-{\bm{\eta }}
    _{t}^{IMU}) ={C^{ - 1}}{\bf{\tilde u}}_{t}^{IMU}+{\bm{\delta }_{t}}\label{eq:omega_calibrated}
  \end{equation}
  where ${\bf{u}}_{t}^{IMU}$ is the calibrated IMU measurements at time $t$;
  ${\bf{\tilde u}}_{t}^{IMU}$ is the raw IMU measurements at time $t$, which can
  be either angular velocity or linear acceleration; ${\bf{b}}_{t}^{IMU}$ and
  ${\bm{\eta }}_{t}^{IMU}$ are the bias and noise of the IMU respectively; ${\bf{\delta }}
  ={C^{ - 1}}({\bf{b}}_{t}^{IMU}+{\bf{\eta }}_{t}^{IMU})$ is defined as the IMU correction
  term, which is a time-varying error.

  To compute the IMU correction term ${\bf{\delta }}_{t}$, the SNN introduced in
  Section IV.C is employed to make predictions based on a local temporal window of size $N$, i.e., ${\bf{u}}_{t-N}^{IMU},\ldots,{\bf{u}}_{t}^{IMU}$.
  A learnable calibration matrix ${\bf{C}}$, initialized as the identity matrix ${{\bf{I}}_3}$, is optimized during training to correct measurement distortions.  This allows the SNN to refine the IMU measurements from the initial training epochs, effectively denoising and debiasing the inputs before feeding them into the InEKF. Once calibrated, the corrected IMU data are integrated into the InEKF framework, which updates the system state $\bf{x_t}$ by incorporating ZUPT constraints based on the robot's kinematic model, i.e., Eq.\ref{eq:angular}-Eq.\ref{eq:position}.
  However, the performance of the InEKF still depends on accurate modeling and predefined noise parameters, which are often difficult to obtain in practice. To address this limitation, the SNN is further utilized to dynamically infer the observation noise covariance matrix $\bf{N}_{t}$ in real-time from the calibrated IMU data. By providing both denoised inputs and adaptive noise estimates, the proposed SNN-InEKF framework enhances the robustness and accuracy of pose estimation, including rotation ${\bf{R}}_{t}^{IMU}$, velocity ${\bf{v}}_{t}
  ^{IMU}$ and postion ${\bf{p}}_{t}^{IMU}$.

  \subsection{InEKF with Pseudo-measurements}
  The IMU dead reckoning can be formulated as a time-series problem, wherein a sequence of IMU measurements collected over a temporal window is used to recursively estimate the system's state. Given an initial state comprising the rotation matrix, velocity, and position, denoted as
  $({\bf{R}}_{0}^{IMU},{\bf{v}}_{0}^{IMU},{\bf{p}}_{0}^{IMU})$, the subsequent IMU readings, i.e., 
  $({\bm{\omega }}_{t}^{IMU},{\bf{a}}_{t}^{IMU})$, can be integrated through the system’s kinematic model to predict the state at the next time step.
  \begin{equation}
    {{\bf{x}}_t}={[\begin{array}{*{20}{c}}{{\bf{R}}_t^{IMU}}&{{\bf{v}}_t^{IMU}}&{{\bf{p}}_t^{IMU}}&{{\bf{b}}_t^\omega }&{{\bf{b}}_t^a}&{{\bf{R}}_t^c}&{{\bf{p}}_t^c}\end{array}]^T}
    \label{eq:state_prediction}
  \end{equation}
  where ${\bf{R}}_{t}^{IMU}$ is the rotation matrix from the world frame to the
  IMU frame at time $t$; ${{\bf{v}}_t^{IMU}}$ and ${{\bf{p}}_t^{IMU}}$ are the velocity
  and position of the mobile robot in the world frame at time $t$;
  ${\bf{b}}_{t}^{\omega}$ and ${\bf{b}}_{t}^{a}$ are the gyroscope and
  accelerometer biases at time $t$; ${\bf{R}}_{t}^{c}$ describes the rotation matrix
  from the mobile robot frame to the IMU frame; ${\bf{p}}_{t}^{c}$ denotes the
  displacement vector from the mobile robot frame to the IMU frame. Thus,
  ${{\bf{x}}_t}$ is also the state vector of the InEKF in our proposed method
  and the state transition function ${\bf{f}}$ can be derivated from the
  kinematic model of the mobile robot, i.e., Eq.\ref{eq:angular}-Eq.\ref{eq:position}.

  Consider the different frames, the velocity of the origin point of the mobile
  robot frame can be expressed as,
  \begin{equation}
    {\bf{v}}_{t}^{c}= \left[{\begin{array}{*{20}{c}}{v_t^{fwd}}\\ {v_t^{lat}}\\ {v_t^{up}}\end{array}}
    \right] ={({\bf{R}}_t^c)^T}({({\bf{R}}_t^{IMU})^T}{\bf{v}}_{t}^{IMU}+{({{\bf{\omega }}_t} - {\bf{b}}_t^\omega )_ \times }
    {\bf{p}}_{t}^{c})\label{eq:velocity_origin}
  \end{equation}
  where ${v_t^{fwd}}$, ${v_t^{lat}}$, and ${v_t^{up}}$ are the forward, lateral,
  and upward velocity of the mobile robot frame, respectively.

  According to the kinematic constraints of the wheeled mobile robot operating
  under nonholonomic conditions, the lateral velocity and vertical velocity should
  theoretically be zero, i.e., $v_{t}^{lat}=0$ and $v_{t}^{up}=0$. Therefore, we
  can use the lateral and upward velocity as pseudo-measurements to update the
  state of the InEKF. The pseudo-measurements can be expressed as,
  \begin{equation}
    {{\bf{y}}_t}= h({{\bf{x}}_t}) = \left[{\begin{array}{*{20}{c}}{y_t^{lat}}\\ {y_t^{up}}\end{array}}
    \right] = \left[{\begin{array}{*{20}{c}}{{h^{lat}}({{\bf{x}}_t}) + n_t^{lat}}\\ {{h^{up}}({{\bf{x}}_t}) + n_t^{up}}\end{array}}
    \right] = \left[{\begin{array}{*{20}{c}}{v_t^{lat}}\\ {v_t^{up}}\end{array}}\right
    ] +{n_t}\label{eq:measurement_pseudo}
  \end{equation}
  where ${\bf{h}}({\bf{x}}_{t})$ is the measurement function that maps the state
  vector to the measurement space, and ${n_t}$ is the measurement noise, which
  is assumed to be Gaussian with zero mean and a certain covariance matrix
  ${\bf{N}}_{t}$. The InEKF is then fed with the pseudo-measurements ${\bf{y}}_{t}$
  and the state vector ${\bf{x}}_{t}$ is updated based on the incoming measurements
  and the system dynamics. Afterwards, the noise covariance matrices ${\bf{N}}_{t}$
  is used to update the InEKF gain, i.e.,
  \begin{equation}
    {{\bf{K}}_t}={{\bf{P}}_t}{{\bf{H}}_t}^{T}({{\bf{H}}_t}{{\bf{P}}_t}{{\bf{H}}_t}
    ^{T}+{{\bf{N}}_t})^{-1}\label{eq:gain_update}
  \end{equation}
  where ${\bf{K}}_{t}$ is the Kalman gain, and ${\bf{H}}_{t}$ is the Jacobian matrix
  of the measurement function ${\bf{h}}$ with respect to the state vector ${\bf{x}}
  _{t}$.

  \subsection{Proposed SNN Architecture}
 SNNs are brain-inspired models that encode and process information through discrete spike events occurring at specific time points. Unlike traditional DNNs, SNNs more closely emulate the information transmission mechanisms observed in biological neural systems, rendering them particularly suitable for neuroscience-inspired modeling and highly compatible with energy-efficient neuromorphic hardware platforms \cite{roy2019towards}. In recent years, SNNs have demonstrated strong potential in handling temporal dynamics and extracting spatial features from time-series data, offering promising capabilities for applications requiring real-time and low-power computation.

  In the proposed method, an SNN is designed to predict both the IMU correction term and the observation noise parameters of the InEKF. Given a temporal window of $N$ inertial measurements as input, the network processes the data to compute the necessary corrections and dynamically adjust the noise model parameters,
  \begin{equation}
    {{\bf{y}}_{net}}= SNN(\{{\bm{\omega }}_{t}^{IMU},{\bf{a}}_{t}^{IMU}\}_{i = t
    - N}^{t})\label{eq:y_net}
  \end{equation}
  where ${\bf{y}}_{net}$ is the output of the SNN, which is a 14-dimensional
  vector containing the IMU correction term ${\bm{\delta }}_{t}$ and the noise
  parameters ${\bf{N}}_{t}$ of the InEKF, i.e.,

  \begin{equation}
    \label{eq13}
    \begin{array}{*{20}{c}}
      {{\bf{y}}_{net}}=[{y^{{c_1}}} & {y^{{c_2}}} & {y^{{c_3}}} & {y^{{c_4}}} & {y^{{c_5}}} & {y^{{c_6}}} & {y^{{b_1}}}   & {y^{{b_2}}} \\
                                    & {y^{{b_3}}} & {y^{{b_4}}} & {y^{{b_5}}} & {y^{{b_6}}} & {y^{{r_1}}} & {y^{{r_2}}}]^T
    \end{array}
  \end{equation}

  Specifically, through the SNN predict value ${\bf{y}}_{net}$, the inverse of calibration
  matrix ${C^{ - 1}}$ of the IMU error simplified model can be obtained as \cite{guo2023model},
  \begin{equation}
    {C^{ - 1}}= diag({10^{\beta {y^{{c_1}}}}},{10^{\beta {y^{{c_2}}}}},{10^{\beta {y^{{c_3}}}}}
    ,{10^{\beta {y^{c_4}}}},{10^{\beta {y^{{c_5}}}}},{10^{\beta {y^{{c_6}}}}})
  \end{equation}
  where $\beta$ is a scaling factor to adjust the range of the calibration
  matrix elements and it is set to 0.1 in this paper according to engineering practice. The bias vector ${\bf{b}}_{t}^{IMU}$ can be dictated as,
  \begin{equation}
    {\bf{b}}_{t}^{IMU}={[{y^{{b_1}}},{y^{{b_2}}},{y^{{b_3}}},{y^{{b_4}}},{y^{{b_5}}},{y^{{b_6}}}]^T}
  \end{equation}.

  For InEKF, the parameters that need to be adjusted including the initial covariance matrix
  ${{\bf{P}}_0}$, the process noise covariance matrix ${\bf{Q}}_{t}$ and the
  measurement noise covariance matrix ${\bf{N}}_{t}$. For ${{\bf{P}}_0}$ and ${\bf{Q}}
  _{t}$ are initialized as a diagonal matrix with small values, which can be
  dynamically adapted for future iterations. Regarding ${\bf{N}}_{t}$, it can be
  estimated as follows,
  \begin{equation}
    {{\bf{N}}_t}= diag(\sigma_{lat}^{2}{10^{{y^{r1}}}},\sigma_{up}^{2}{10^{{y^{r2}}}}
    )
  \end{equation}
  where $\sigma_{lat}^{2}$ and $\sigma_{up}^{2}$ are the initial covariance values
  of the lateral and upward velocity, respectively. ${{\bf{N}}_t}$ may inflate covariance
  up to a factor $10^{{y^{r1}}}$ and squeeze it up to a factor $10^{{y^{r2}}}$.
  Given that wheeled mobile robots are susceptible to lateral slip and other
  conditions that can compromise the accuracy of the ZUPT method, real-time adjustments
  to ${\bf{N}}_{t}$ are essential. To address this, our approach dynamically adapts
  ${\bf{N}}_{t}$ based on the SNN output. This adaptive strategy mitigates the
  interference from pseudo-observations with lower reliability, thereby enhancing
  the positioning accuracy of the InEKF algorithm.

\begin{figure*}[!ht]
    \centering
    \includegraphics[width=.99\linewidth]{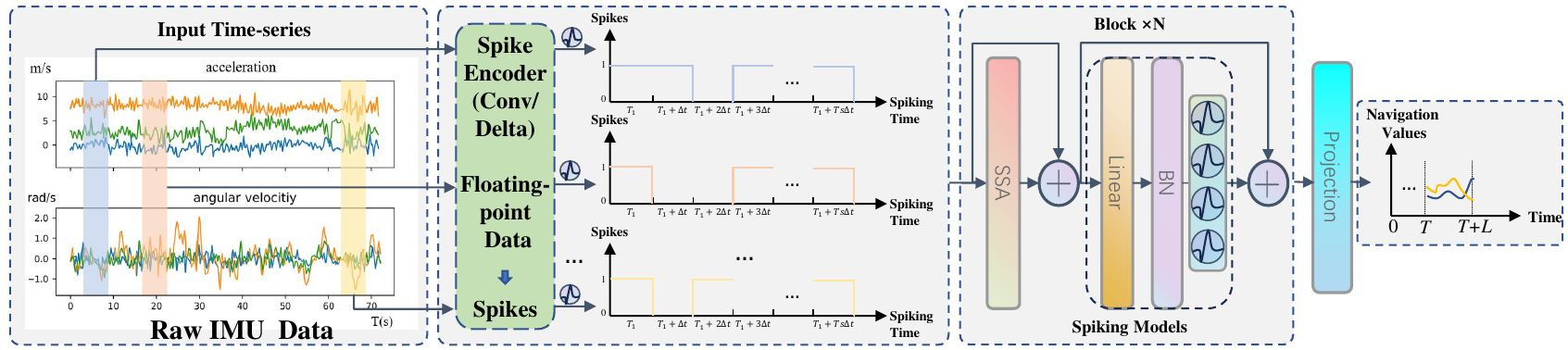}
    \caption{The pipeline of SNN for IMU dead reckoning. }
    \label{fig:spike-snn}
  \end{figure*}
  
  To fully leverage the inherent characteristics of SNNs, it is essential to
  align the temporal dimension between IMU time-series data and SNNs. As shown in Fig. \ref{fig:spike-snn}, 
  the core strategy in the paper is that revolves around integrating the relevant finer details of
  spike activity within the time-series data at each time step. To achieve this,
  a time step $\Delta T$ of the time series is divided into Ts segments. Each of
  these segments enables a firing event for neurons whose membrane potentials
  exceed the threshold, thereby ensuring $\Delta T$ is equivalent to $\Delta T ={T_s}
  \Delta t$. This equation serves as a bridge between the time step $\Delta T$
  in a time series and the time step $\Delta t$ in the SNN. Consequently, the
  independent variable $t$, as employed in both the time-series context (denoted
  as $X(t)$) and within the SNN framework $(U(t), I(t), H(t), S(t))$, now holds
  a shared significance. In a recent study, Qu et al. \cite{qu2024cnn} compellingly
  demonstrated that this specific type of morphological information can be effectively
  modeled employing a distinct variety of CNN kernels. Building on this
  insightful finding, the innovative utilization of a convolutional layer is selected
  as an optimal temporal encoder. Given the $i_{th}$ historical raw IMU data
  $\bf{U}_{i}=({\bf{u}}_{t-N}^{IMU},\ldots,{\bf{u}}_{t}^{IMU})$, it can be input
  to the convolutional layer followed by batch normalization and generate the spike
  as,
  \begin{equation}
    {\bf{S}}= SN(BN(Conv({{\bf{U}}_i})))\label{eq:spike_input}
  \end{equation}

  By passing through the spike encoder based on the convolutional layer, the
  dimension of ${{\bf{U}}_i}$ can be expanded to ${T_s}\times T \times C$, where
  $C$ is the number of channels, i.e. 6, the 3-axis angular velocity, and the 3-axis
  linear acceleration. The convolutional spike encoder effectively captures the internal
  temporal information inherent in the input data, namely, the temporal changes and
  shapes, which significantly contributes to the robust representation of the
  dynamic nature of this information over time. This encoding process adeptly caters
  to the subsequent spiking layers, thereby facilitating event-driven modeling.

  Considering the high sampling frequency of the IMU, usually 100Hz, and the
  presence of substantial nonlinear noise components in IMU data, the SNN needs to
  learn the temporal features of the IMU measurements over a long time window.
  Traditional neural networks, such as RNN or CNN, often struggle to capture long-term
  dependencies due to the vanishing gradient problem. To address this issue, we employ
  a spiking Transformer as the backbone of the SNN, which is capable of
  capturing long-term dependencies and temporal features in the IMU measurements.
  In the proposed SNN architecture, the spiking version of Transformer is built
  upon the iTransformer \cite{liu2023itransformer}, which is the state-of-the-art
  time-series forecasting model on several benchmark datasets.

  Inspired by Spikformer v2 \cite{zhou2022spikformer}, the spiking self-attention
  (SSA) mechanism is incorporated into the iTransformer architecture to build
  the spiking Transformer blocks. The SSA mechanism is designed to capture the temporal
  dependencies and long-term relationships in the input data, which is essential
  for accurately modeling the IMU measurements. Specifically, a channel-wise
  spiking embedding layer is used to convert the input spike trains into a
  continuous representation, which is then fed into the spiking Transformer blocks
  and can be expressed as follows,
  \begin{equation}
    {{\bf{S}}_{emb}}= SN(Linear({\bf{S}}))\label{eq:spiking_embedding}
  \end{equation}
  where ${\bf{S}}_{emb}$ is the spiking embedding of the input spike trains
  ${\bf{S}}$, and the linear layer is used to project the spike trains into a higher-dimensional
  space.

  The spiking Transformer blocks consist of multiple layers of spiking self-attention
  and feed-forward networks, which are designed to capture the temporal features
  and long-term dependencies in the input data. The output of the spiking Transformer
  blocks are then fed into a projection layer, such as a fully connected layer, to
  generate the final output of the SNN, i.e., the IMU correction term ${\bm{\delta }}
  _{t}$ and the noise parameters ${\bf{N}}_{t}$ of the InEKF.

  The proposed SNN architecture is trained using backpropagation through time
  (BPTT) \cite{gruslys2016memory} algorithm, which is a common training method
  for SNNs. In the proposed SNN optimization framework, the relative
  displacement $\Delta{\bf{p}}$, the alteration in relative velocity $\Delta{\bf{v}}$,
  and the relative rotation $\Delta{\bf{R}}$ at each instant are designated as
  the optimization variables. The comprehensive training loss is subsequently derived
  from the Huber loss function \cite{meyer2021alternative}, which helps in
  balancing robustness and sensitivity to outliers during the training process. The
  loss function can be expressed as,
  \begin{equation}
    L_{\text{Huber }}(y, \hat y)=\left\{
    \begin{array}{ll}
      \frac{1}{2}(y-\hat y)^{2},               & |y-\hat y| \leq \delta \\
      \delta|y-\hat y|-\frac{1}{2} \delta^{2}, & |y-\hat y|>\delta
    \end{array}\right.
  \end{equation}
  where the Huber parameter $\delta$ is set $4e-4$. The loss function is constructed to be fully differentiable with respect to the parameters of the SNN, thereby enabling efficient optimization via gradient-based learning algorithms. During training, the SNN parameters are iteratively updated to minimize the loss, which in turn enhances the accuracy of the IMU dead reckoning estimates by improving the quality of the predicted correction terms and noise parameters.

  \section{Experimental Results and Analysis}
  \subsection{Dataset and Evaluation Metrics}
  To evaluate the performance of the proposed method, the KITTI odometry dataset
  \cite{geiger2012we} is used, which serves as a widely adopted benchmark for
  evaluating visual-inertial odometry and SLAM algorithms. The KITTI dataset contains
  a series of sequences captured by a stereo camera and an IMU mounted on a vehicle,
  which provides ground truth poses for evaluation. The dataset consists of 22 sequences,
  each with a different length and varying environmental conditions. We download
  the raw data with the IMU measurements sampled at 100 Hz. In this article, two
  widely recognized evaluation metrics are employed to assess the accuracy of the
  proposed method, which are also used in the KITTI dataset, i.e., KITTI odometry
  evaluation metrics,

  1) Relative Translational Error (RTE): It computes across all possible trajectory
  subsequences with lengths ranging from 100 to 800 meters, expressed as a
  percentage of the total traversed distance, and can be defined as follows,
  \begin{equation}
    RTE=\frac{1}{|\mathcal{F}|}\sum_{(i, j) \in \mathcal{F}}\frac{\operatorname{Trans}\left(\left(\hat{p}_{j}\ominus
    \hat{p}_{i}\right) \ominus\left(p_{j}\ominus p_{i}\right)\right)}{\operatorname{length}(i,
    j)}\label{eq:RTE}
  \end{equation}
  where $\hat{p}_{i}$ and $p_{i}$ are the estimated and ground truth poses at
  time $i$, respectively; $\ominus$ denotes the operation of transforming the
  pose from the world frame to the IMU frame; $\operatorname{Trans}$ represents that
  computes the translation value of the pose; $\operatorname{length}(i, j)$ is
  the length of the subsequence from time $i$ to time $j$, i.e., ranging from 100
  to 800 meters.

  2) Relative Rotation Error (RRE): It represents the relative rotational increment
  error for all potential subsequences ranging in length from 100 to 800 meters,
  expressed in degrees per kilometer, and can be defined as follows,
  \begin{equation}
    RRE=\frac{1}{|\mathcal{F}|}\sum_{(i, j) \in \mathcal{F}}\frac{\operatorname{Rot}\left(\left(\hat{p}_{j}\ominus
    \hat{p}_{i}\right) \ominus\left(p_{j}\ominus p_{i}\right)\right)}{\operatorname{length}(i,
    j)}\label{eq:RRE}
  \end{equation}
  where $\operatorname{Rot}$ represents that computes the rotation value of the
  pose.

  \subsection{Experimental Setup}
  The KITTI raw dataset is acquired utilizing the RT 3000v3 IMU produced by OxTS,
  renowned for its superior accuracy but at a high cost \cite{geiger2012we}. To
  generate low-cost IMU data, a predefined quantity of noise and bias is intentionally
  introduced into the raw KITTI IMU data. Specifically, gyroscope measurements are
  corrupted with Gaussian noise $N(0,{10^{ - 3}})$ and a random bias $U(0.015,0.0
  25)$, whereas accelerometer data is augmented with Gaussian noise
  $N(0,{10^{ - 2}})$ and a bias $U(0.45,0.55)$.

  A total of 14 sequences (Seq. 1–14), each exceeding 80 seconds in duration, are selected for evaluation. Among them, the first 40 seconds of Seq. 6-14 are used as the training set, while the remaining portions of these sequences are reserved for validation. Seq. 1-5 are designated as the test set, where GNSS data serve as ground-truth references for quantitative assessment. To address the mismatch in sampling frequencies between the GNSS and IMU data, linear interpolation is employed to resample the ground-truth trajectories to match the IMU’s 100 Hz rate. All experiments are initialized from the current robot state, and the proposed method is benchmarked against three representative baseline approaches.

  \begin{itemize}
    \item ORB-SLAM3-Stereo \cite{campos2021orb}: An advanced visual SLAM system
      that utilizes stereo cameras for accurate pose estimation without the
      loopclosing.

    \item AI-IMU \cite{brossard2020ai}: A deep learning-based method that
      employs a neural network to estimate the pose of the mobile robot using
      IMU data.

    \item MD-IMU \cite{guo2023model}: A model-driven method that combines a
      Kalman filter with a neural network to estimate the pose of the mobile
      robot using IMU data.
  \end{itemize}

  The proposed method is implemented using PyTorch and SpikingJelly
  \cite{fang2023spikingjelly}. To reduce the risk of overfitting and accelerate
  the training process, the SNN is trained using an Adam optimizer
  \cite{kingma2014adam} with a learning rate initialized $1e-4$ with cosines
  warming restart scheduler \cite{loshchilov2016sgdr}. The weight decay is set to
  $5e-2$, and the dropout rate is set to $0.1$. In addition, training is
  conducted for 1000 epochs. To capture more temporal information, the SNN is
  designed with a window size of $N=500$, which means that the SNN takes the last
  500 IMU measurements as input to predict the IMU correction term and the noise
  parameters of the InEKF. The dimension of the embedding layer and feed-forward
  layer in the Spike-Transformer is set to 256, and the number of heads in the self-attention
  mechanism is set to 4. The number of spiking Transformer blocks is set to 2,
  and the number of neurons in the output layer is set to 14, which corresponds to
  the IMU correction term and the noise parameters of the InEKF. The proposed
  SNN is trained on a single NVIDIA RTX A6000 GPU with a batch size of 8.

  \subsection{Experimental Results on the KITTI Dataset}
\begin{figure}[!ht]
    \centering
    \includegraphics[width=.99\linewidth]{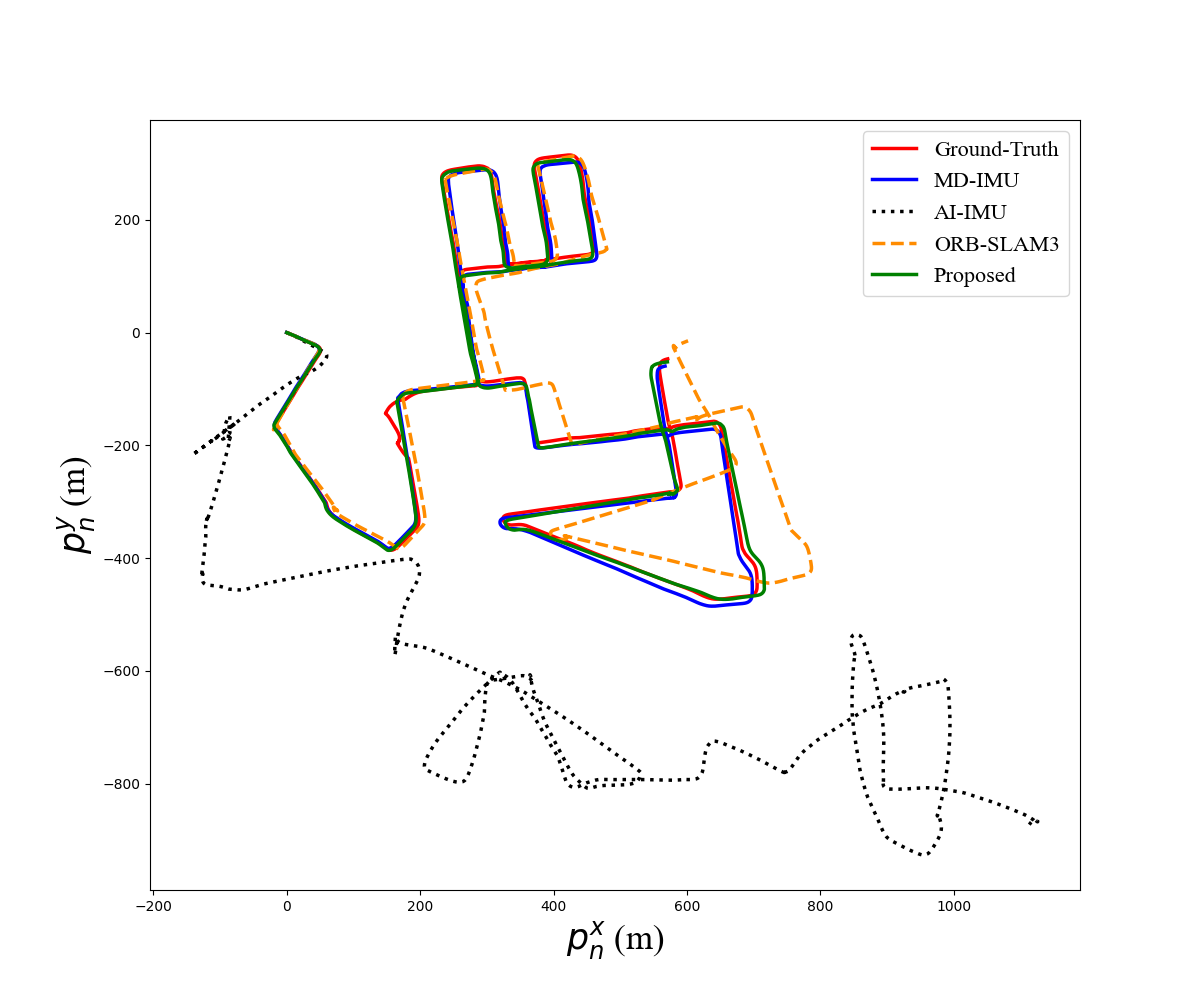}
    \caption{Test results on Seq.3 in the KITTI dataset. }
    \label{fig:Seq3_result}
  \end{figure}

  The experimental results of the proposed method and baseline approaches on the KITTI dataset are shown in Fig. \ref{fig:Seq3_result} and Tab. \ref{tab::kitti_result}, evaluated in terms of RTE, RRE, and per-sequence processing time. As illustrated in Fig. \ref{fig:Seq3_result}, the trajectory of Seq. 3 estimated by MD-IMU, AI-IMU, ORB-SLAM3, and the proposed method is compared against the ground truth, represented by blue, black, orange, and green curves, respectively. The proposed method demonstrates superior accuracy over MD-IMU and AI-IMU across both RTE and RRE metrics. This improvement stems from the limited ability of MD-IMU and AI-IMU to suppress the substantial noise present in low-cost IMU data, with MD-IMU further constrained by its reliance on a short time window for InEKF noise parameter estimation. In contrast, the proposed method utilizes a spike encoder to transform raw IMU measurements into spike trains, which are then processed by spiking Transformer blocks to extract temporal features and long-range dependencies. This mechanism enables the denoised IMU signals to more accurately reflect the true motion dynamics of the robot, thereby enhancing the performance of InEKF and improving localization accuracy. On average, the proposed method achieves lower RTE than ORB-SLAM3 on Seq. 12 and 13, though it underperforms slightly in RRE. It is worth noting that the localization performance of ORB-SLAM3 Stereo could be further improved with loop-closure enabled, which was not activated in this evaluation.

\begin{figure}[!ht]
    \centering
    \includegraphics[width=.9\linewidth]{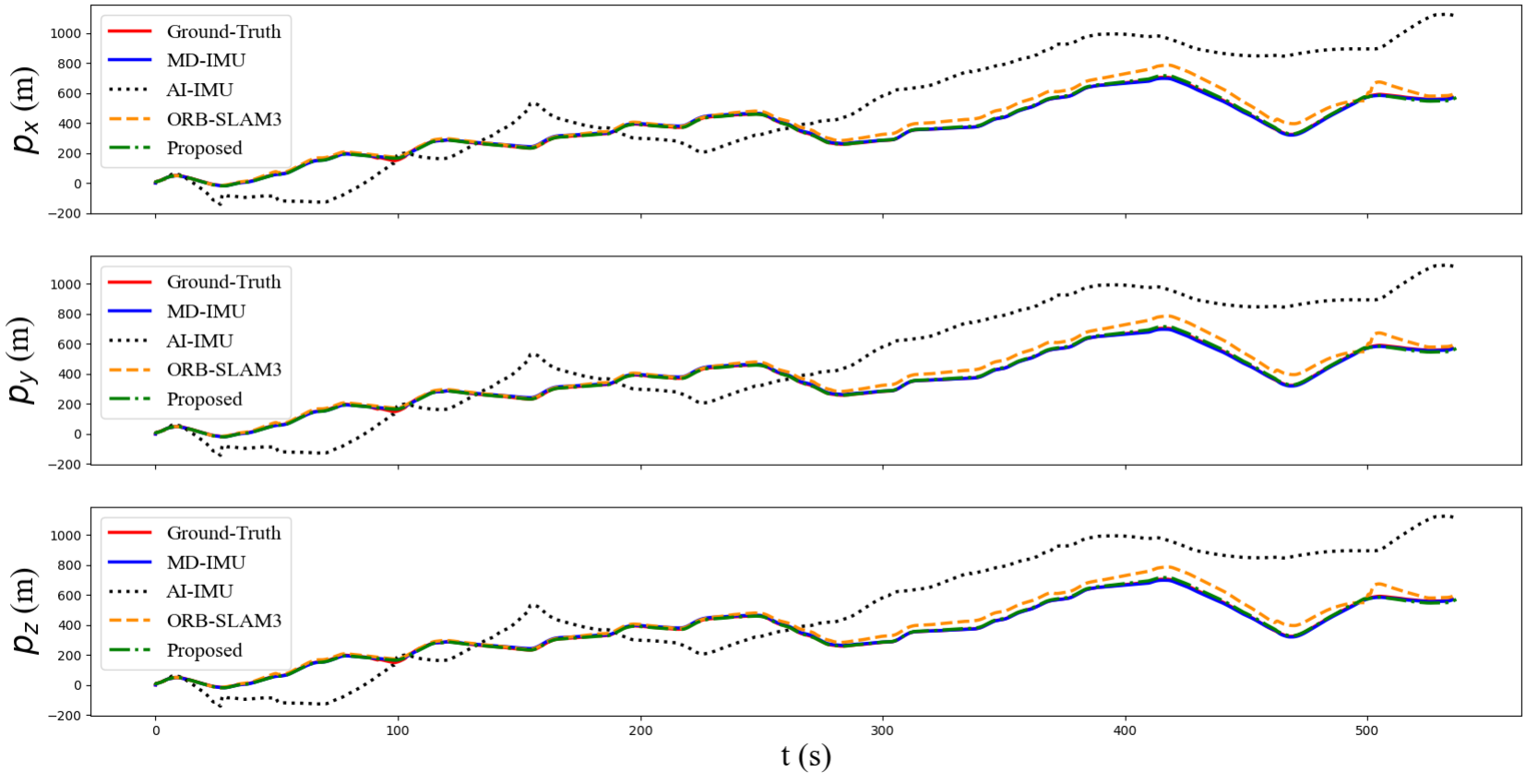}
    \caption{The position results on Seq.3 in the KITTI dataset. }
    \label{fig:Seq3_p}
  \end{figure}

\begin{figure}[!ht]
    \centering
    \includegraphics[width=.9\linewidth]{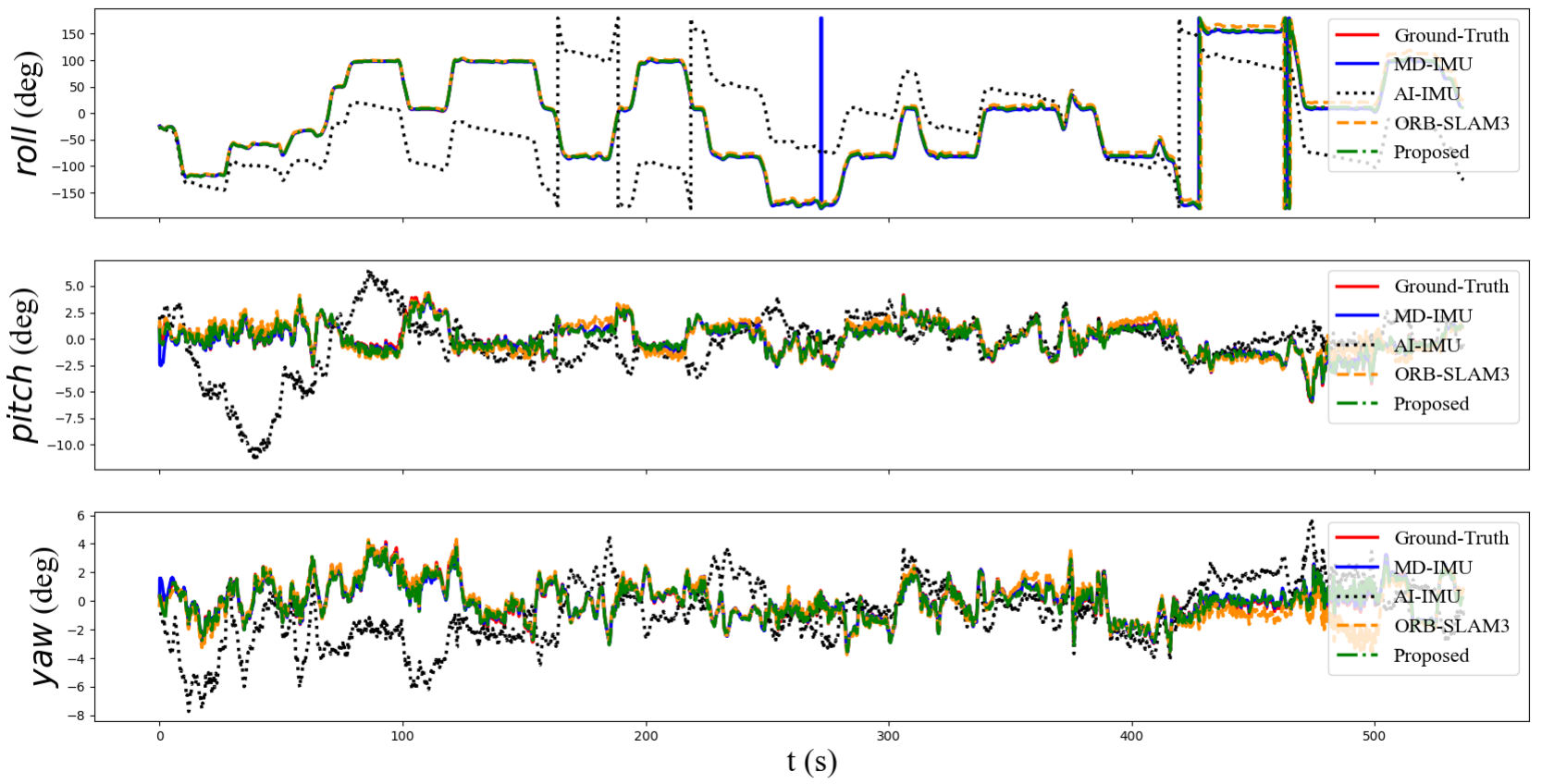}
    \caption{The attitude results on Seq.3 in the KITTI dataset. }
    \label{fig:Seq3_a}
  \end{figure}

  The position and attitude estimations are further illustrated in Fig. \ref{fig:Seq3_p} and Fig. \ref{fig:Seq3_a}, respectively. The AI-IMU method exhibits significant drift in both translational and rotational estimates, while the proposed method shows high fidelity to the ground truth, demonstrating its efficacy in mitigating the effects of noise and bias in low-cost MEMS IMUs. These results confirm that the proposed approach is competitive with, and in several cases outperforms, state-of-the-art methods such as ORB-SLAM3 and MD-IMU.

  \begin{table*}
    [ht]
    \centering
    \caption{ Experimental results of different methods on the KITTI Dataset}
    \label{tab::kitti_result}
    \scalebox{0.9}{
    \begin{tabular}{ccc|ccc|ccc|ccc|ccc}
      \hline
      \toprule[1pt] \multicolumn{1}{c|}{\multirow{2}{*}{Seq.}} & \multicolumn{1}{c|}{\multirow{2}{*}{Sequence Name}} & \multirow{2}{*}{Length(m)} & \multicolumn{3}{c|}{ORB-SLAM3} & \multicolumn{3}{c|}{AI-IMU} & \multicolumn{3}{c|}{MD-IMU} & \multicolumn{3}{c}{Proposed} \\
      \cline{4-15} \multicolumn{1}{c|}{}                       & \multicolumn{1}{c|}{}                               &                            & RTE                                   & RRE                         & T(s)                        & RTE                         & RRE   & T(s)  & RTE  & RRE          & T(s)         & RTE           & RRE           & T(s)  \\
      \hline
      \multicolumn{1}{c|}{1}                                   & \multicolumn{1}{c|}{2011\_09\_26\_drive\_0036}      & 743.9                      & 4.8                                   & 19.1                        & 41.5                        & 115.4                       & 188.2 & 15.5  & 2.9  & 3.8          & 31.9         & \textbf{1.4}  & \textbf{2.5}  & 40.2  \\
      \multicolumn{1}{c|}{2}                                   & \multicolumn{1}{c|}{2011\_09\_26\_drive\_0101}      & 1312.6                     & \textbf{3.3}                          & 4.9                         & 51.8                        & 1273.1                      & 35.2  & 19.9  & 3.6  & 3.3          & 30.4         & 3.5           & \textbf{0.8}  & 42.3  \\
      \multicolumn{1}{c|}{3}                                   & \multicolumn{1}{c|}{2011\_09\_30\_drive\_0028}      & 4243.4                     & 5.9                                   & 24.3                        & 81.0                        & 102.4                       & 152.4 & 33.8  & 9.1  & 5.2          & 49.8         & \textbf{2.6}  & \textbf{2.5}  & 60.7  \\
      \multicolumn{1}{c|}{4}                                   & \multicolumn{1}{c|}{2011\_09\_30\_drive\_0034}      & 922.3                      & 3.2                                   & 27.9                        & 63.7                        & 46.3                        & 47.2  & 26.0  & 2.8  & 5.5          & 40.3         & \textbf{2.3}  & \textbf{2.5}  & 52.4  \\
      \multicolumn{1}{c|}{5}                                   & \multicolumn{1}{c|}{2011\_10\_03\_drive\_0042}      & 2600.4                     & 18.1                                  & 19.2                        & 79.4                        & 828.1                       & 14.1  & 27.8  & 6.1  & 2.5          & 35.7         & \textbf{5.3}  & \textbf{1.4}  & 48.5  \\
      \multicolumn{1}{c|}{6}                                   & \multicolumn{1}{c|}{2011\_09\_26\_drive\_0022}      & 522.9                      & 4.7                                   & 42.1                        & 38.3                        & 62.9                        & 85.2  & 16.6  & 1.9  & 6.3          & 22.9         & \textbf{1.6 } & \textbf{4.0}  & 32.3  \\
      \multicolumn{1}{c|}{7}                                   & \multicolumn{1}{c|}{2011\_09\_29\_drive\_0071}      & 245.2                      & 4.5                                   & 11.2                        & 48.3                        & 126.9                       & 767.2 & 22.1  & 4.9  & 6.5          & 31.4         & \textbf{4.0}  & \textbf{2.6}  & 42.8  \\
      \multicolumn{1}{c|}{8}                                   & \multicolumn{1}{c|}{2011\_09\_30\_drive\_0018}      & 2212.6                     & 3.9                                   & 23.2                        & 149.2                       & 85.6                        & 184.1 & 73.7  & 19.2 & 25.5         & 76.9         & \textbf{3.2}  & \textbf{8.2}  & 90.5  \\
      \multicolumn{1}{c|}{9}                                   & \multicolumn{1}{c|}{2011\_09\_30\_drive\_0020}      & 1242.8                     & 4.8                                   & 32.5                        & 57.8                        & 147.1                       & 114.5 & 22.6  & 3.7  & 4.0          & 29.0         & \textbf{2.2}  & \textbf{2.1}  & 42.4  \\
      \multicolumn{1}{c|}{10}                                  & \multicolumn{1}{c|}{2011\_09\_30\_drive\_0027}      & 692.9                      & 4.5                                   & 37.2                        & 51.9                        & 115.8                       & 286.9 & 21.3  & 1.8  & 5.9          & 28.5         & \textbf{1.4}  & \textbf{3.2}  & 40.7  \\
      \multicolumn{1}{c|}{11}                                  & \multicolumn{1}{c|}{2011\_09\_30\_drive\_0033}      & 1714.5                     & 8.1                                   & 27.3                        & 276.2                       & 135.3                       & 178.2 & 185.0 & 2.6  & 5.2          & 134.7        & \textbf{1.8}  & \textbf{2.4}  & 149.4 \\
      \multicolumn{1}{c|}{12}                                  & \multicolumn{1}{c|}{2011\_10\_03\_drive\_0027}      & 3731.8                     & \textbf{3.1}                          & 33.2                        & 244.6                       & 105.2                       & 77.7  & 160.2 & 63.8 & 51.4         & 114.7        & 7.1           & \textbf{23.6} & 128.1 \\
      \multicolumn{1}{c|}{13}                                  & \multicolumn{1}{c|}{2011\_10\_03\_drive\_0034}      & 5087.4                     & \textbf{3.8}                          & 26.1                        & 264.0                       & 119.3                       & 63.9  & 170.3 & 7.1  & 8.2          & 125.0        & 4.2           & \textbf{3.9}  & 139.8 \\
      \multicolumn{1}{c|}{14}                                  & \multicolumn{1}{c|}{2011\_10\_03\_drive\_0047}      & 714.2                      & 10.7                                  & 16.2                        & 44.6                        & 419.1                       & 35.3  & 19.0  & 2.5  & 2.3          & 21.3         & \textbf{1.1}  & \textbf{1.7}  & 36.3  \\
      \hline
      \multicolumn{3}{c|}{Average scores}                      & 6.0                                                 & 24.6                       & /                                     & 263.0                       & 159.3                       & /                           & 9.4   & 9.7   & /    & \textbf{3.0} & \textbf{4.4} & /              \\
      \bottomrule[1pt]
    \end{tabular}
    }
  \end{table*}

\begin{figure}[!ht]
    \centering
    \includegraphics[width=.9\linewidth]{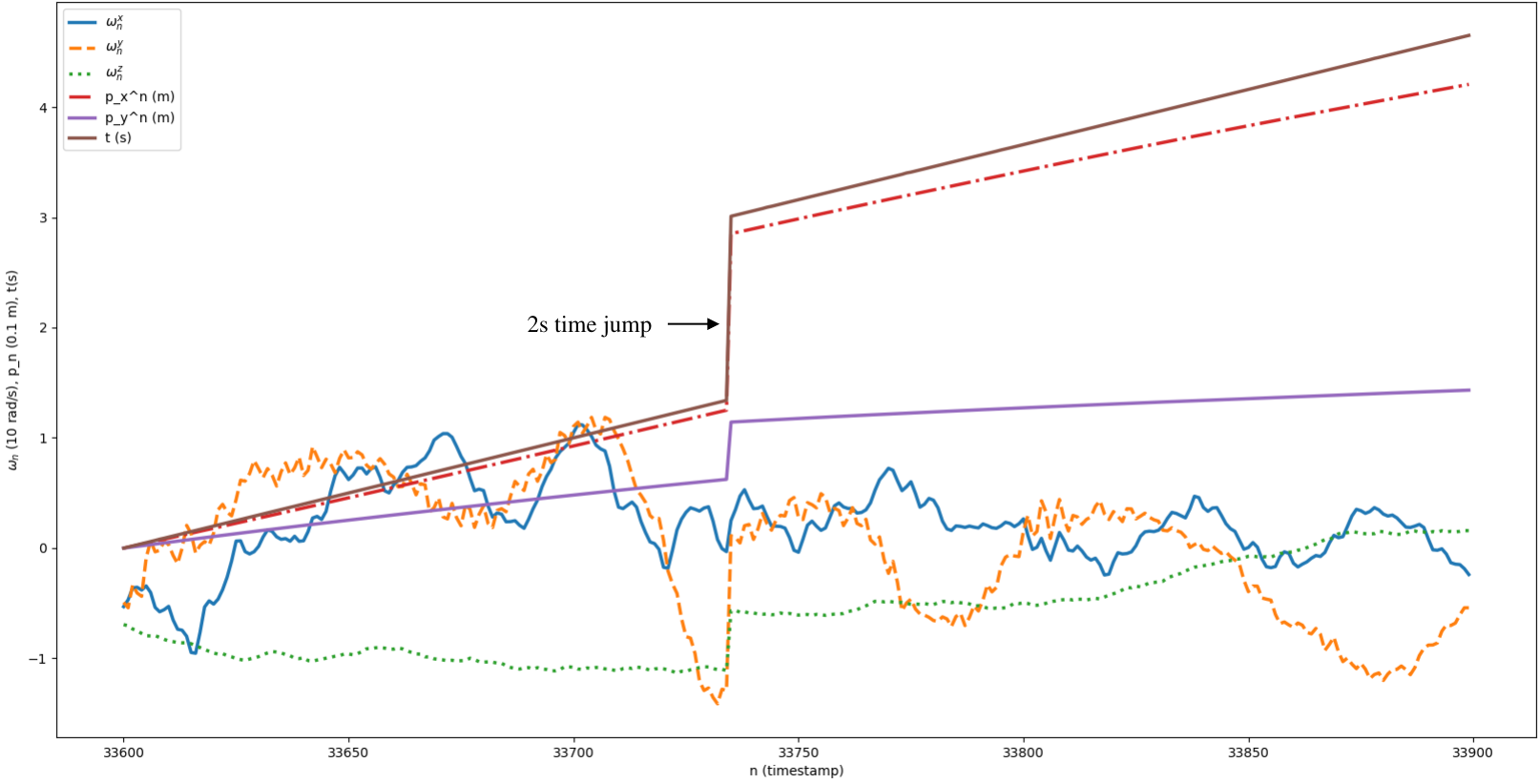}
    \caption{IMU data loss of the Seq.13. }
    \label{fig:Seq13_loss}
  \end{figure}

\begin{figure}[!ht]
    \centering
    \includegraphics[width=.9\linewidth]{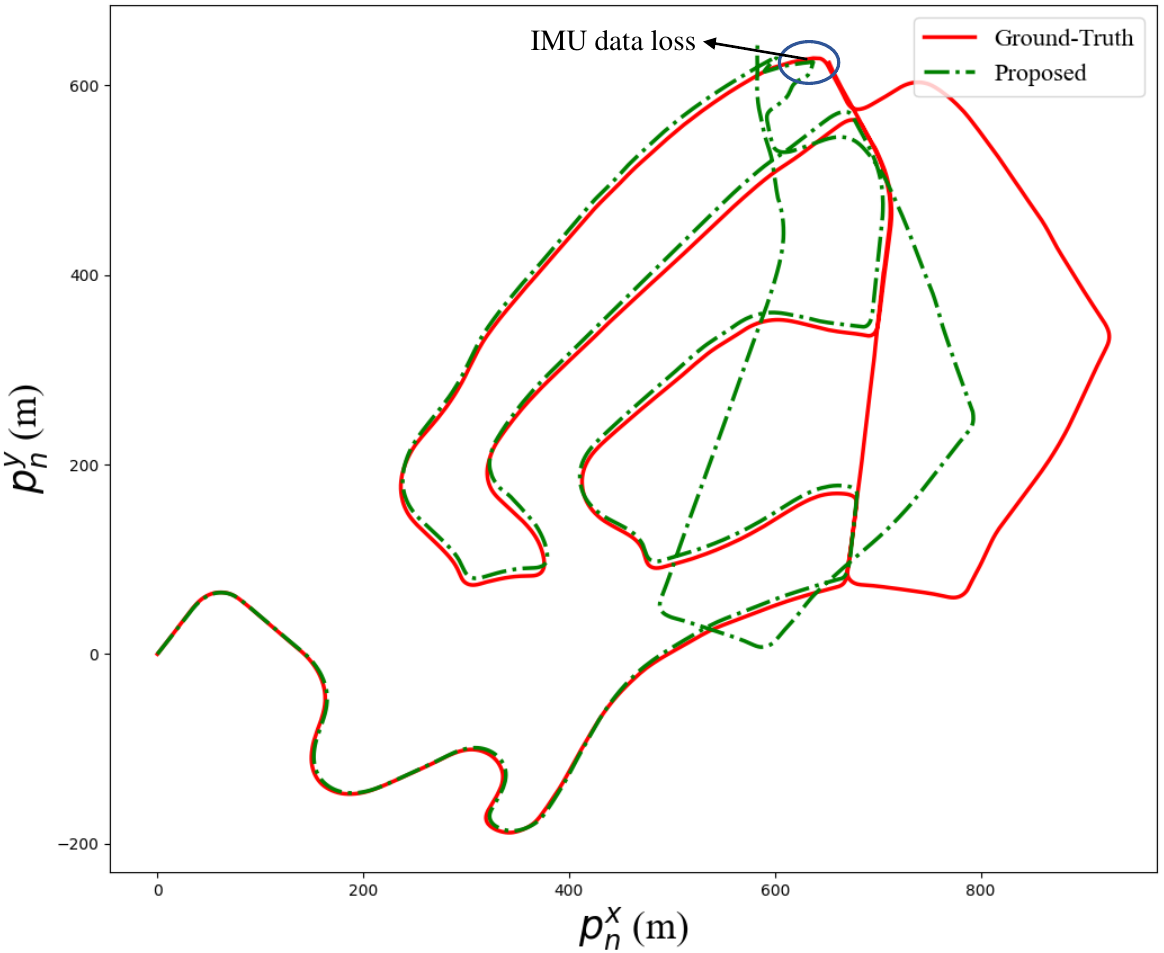}
    \caption{The results of the proposed method on Seq.13 with IMU data loss. }
    \label{fig:Seq13_result}
  \end{figure}

 Seq. 8, 12, and 13 in the KITTI dataset exhibit missing IMU data spanning several seconds, as illustrated in Fig. \ref{fig:Seq13_loss}. Specifically, a 2-second data gap occurs around sample index $n=33750$, resulting in notably increased RTE and RRE values for the MD-IMU methods. In contrast, the proposed method effectively manages this data loss by leveraging available IMU measurements to estimate both the IMU correction term and the noise parameters of the InEKF, as evidenced in Fig. \ref{fig:Seq13_result}. For Seq. 13, the proposed approach reduces the RTE and RRE by 40.8 $\%$ and 52.4 $\%$, respectively, compared to the MD-IMU method. These results underscore the robustness of the proposed method in handling incomplete IMU data while maintaining accurate localization performance. On average, the proposed method achieves RTE and RRE values of 3.0 $\%$ and 4.4 $\deg/km$, significantly outperforming the MD-IMU method, which records 9.7 $\%$ and 9.4 $\deg/km$, respectively.
 
  Tab. \ref{tab::kitti_result} presents a comparative evaluation of computational efficiency across the tested methods, with runtime (seconds) as the performance metric. The ORB-SLAM3 method exhibits the longest runtime due to its reliance on visual data processing. In contrast, the AI-IMU method achieves the shortest runtime, owing to its lightweight network architecture, though it struggles to maintain robust performance when processing data from low-cost IMUs with substantial sensor noise. Notably, the proposed method significantly improves dead-reckoning accuracy compared to the MD-IMU baseline while incurring only a marginal increase in computational time. Furthermore, when deployed on a 45 nm neuromorphic hardware platform \cite{horowitz20141}, the proposed spiking Transformer architecture achieves a theoretical energy consumption reduction of 66.3 $\%$ relative to conventional Transformer-based neural networks \cite{lv2024efficient}. This substantial improvement highlights the energy efficiency of the bio-inspired architecture, making it particularly suitable for edge computing applications with stringent power constraints.

  \subsection{Field Tests}
\begin{figure}[!ht]
    \centering
    \includegraphics[width=.7\linewidth]{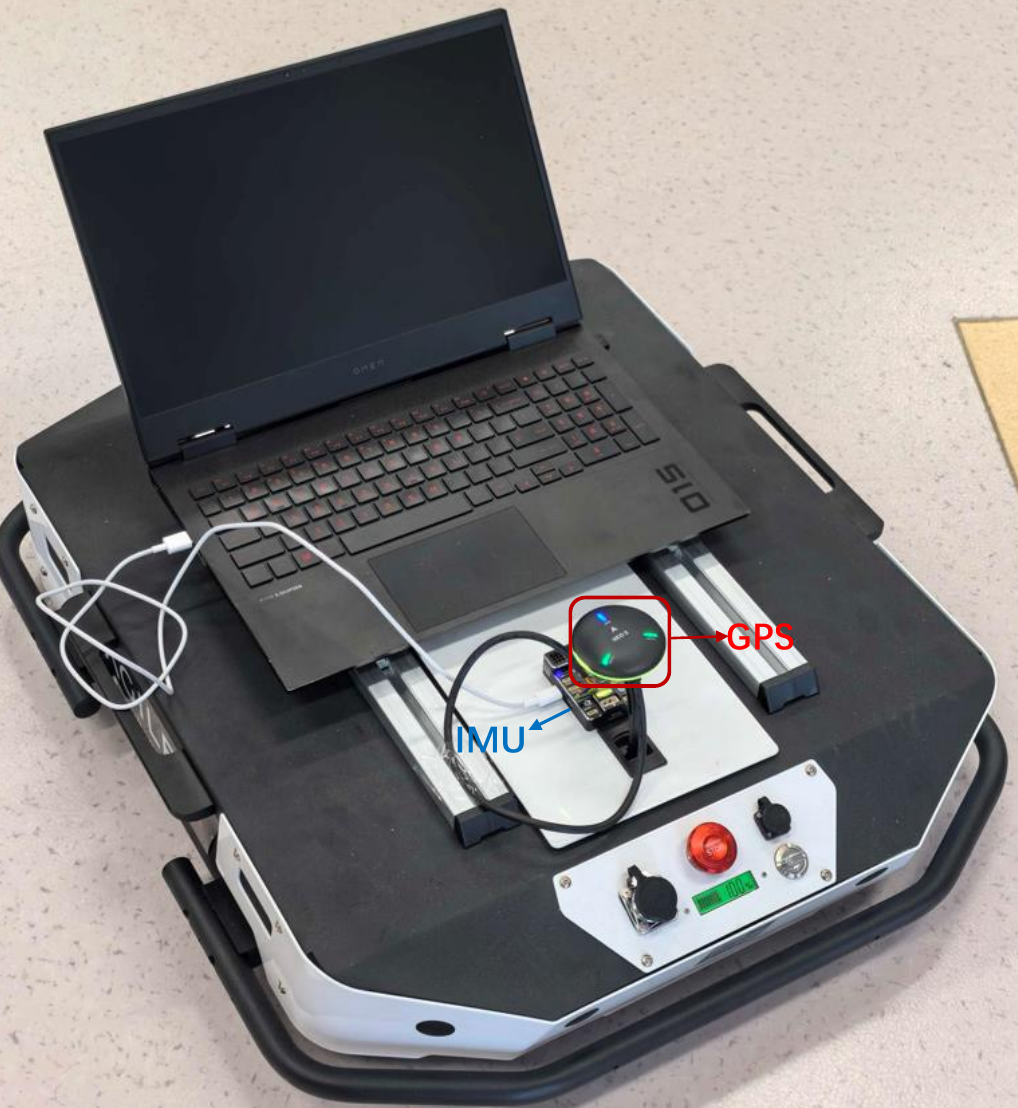}
    \caption{The self-build mobile robot platform. }
    \label{fig:robot}
  \end{figure}

  To further validate the effectiveness of the proposed method, real-world field tests are conducted using a self-developed mobile robot, as illustrated in Fig. \ref{fig:robot}. The robot is equipped with a CUAV Pixhawk V5 Nano autopilot, a high-performance autopilot module suitable for fixed-wing aircraft, multi-rotors, ground vehicles, and other robotic platforms. Ground-truth trajectory data are provided by a Ublox NEO-M9N GPS receiver, offering a horizontal positioning accuracy of approximately 0.7 meters. The proposed method utilizes raw inertial data collected from an ICM-20602 IMU at a sampling rate of 100 Hz. It is worth noting that the ICM-20602 is a low-cost MEMS IMU with substantially lower accuracy compared to the high-precision OxTS RT3000v3 IMU used in the KITTI dataset, highlighting the practical applicability of the proposed approach in resource-constrained scenarios.

  \begin{table}[]
    \centering
    \caption{ Experimental results in the real filed tests}
    \label{tab::field_test}
    \begin{tabular}{cc|cc}
      \hline
      \toprule[1pt] \multicolumn{1}{c|}{\multirow{2}{*}{Seq.}} & \multirow{2}{*}{Length(m)} & \multicolumn{2}{c}{Proposed} \\
      \cline{3-4} \multicolumn{1}{c|}{}                        &                            & RTE                         & RRE  \\
      \hline
      \multicolumn{1}{c|}{1}                                   & 182.1                      & 4.3                         & 4.1  \\
      \multicolumn{1}{c|}{2}                                   & 110.6                      & 6.7                         & 6.4  \\
      \multicolumn{1}{c|}{3}                                   & 99.7                       & 6.1                         & 8.5  \\
      \multicolumn{1}{c|}{4}                                   & 121.3                      & 7.6                         & 10.3 \\
      \multicolumn{1}{c|}{5}                                   & 91.5                       & 5.8                         & 7.2  \\
      \hline
      \multicolumn{2}{c|}{Average scores}                      & 6.1                        & 7.3                          \\
      \hline
      \toprule[1pt]
    \end{tabular}
  \end{table}

  \begin{figure}[!ht]
    \centering
    \includegraphics[width=.9\linewidth]{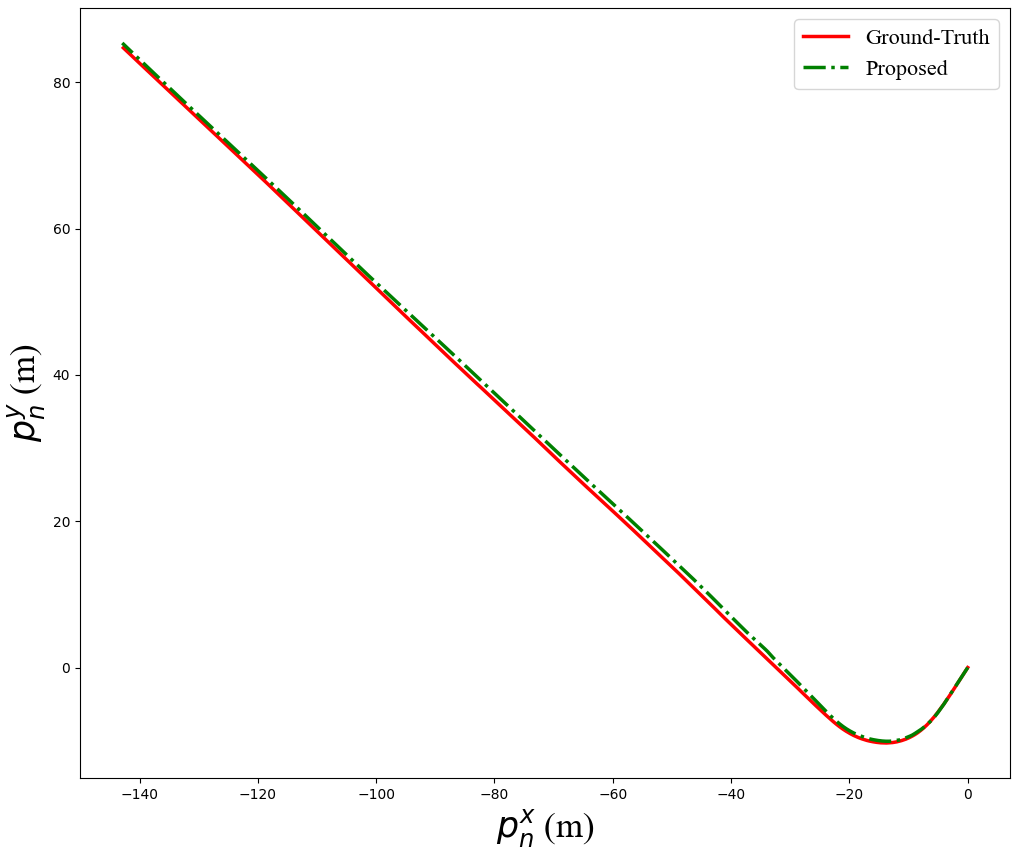}
    \caption{The results on Seq.1 in the real field tests. }
    \label{fig:real_result}
  \end{figure}

  The experimental results of the proposed method on real-world field tests are presented in Tab. \ref{tab::field_test} and Fig. \ref{fig:real_result}. As shown, the proposed method achieves average RTE and RRE values of 6.1 $\%$ and 7.3 $\deg/km$, respectively. It is important to note that the low-cost ICM-20602 IMU used in these tests is highly susceptible to road surface irregularities and chassis vibrations due to its limited design precision. Compared to the high-grade sensor used in the KITTI dataset, this IMU produces significantly noisier measurements under dynamic conditions. In challenging sequences (e.g., Seqs. 3, 4, and 5), where conventional neural networks struggle to optimize parameters due to excessive noise, the proposed method exhibits strong adaptability through cross-sequence feature learning. Specifically, the spiking Transformer architecture effectively captures motion dynamics from higher-quality sequences while modeling noise characteristics via event-driven sparse representations. Although the localization accuracy is slightly reduced compared to results on the KITTI dataset, the proposed method consistently mitigates the effects of noise and bias. As shown in Fig. \ref{fig:real_result}, the estimated trajectory closely aligns with the ground truth, demonstrating the method’s robustness and effectiveness in real-world applications.

  \section{Conclusion}
  In this paper, we propose a novel SNN-based IMU dead reckoning method for low-cost
  IMU. The proposed method employs a spiking Transformer to extract the temporal
  features and long-term dependencies in the IMU measurements, which enables the
  SNN to learn the complex dynamics of the IMU measurements and accurately
  estimate the IMU correction term and the noise parameters of the InEKF. The proposed
  method is evaluated on the KITTI dataset, and the experimental results
  demonstrate that the proposed method outperforms the state-of-the-art methods
  in terms of localization accuracy and robustness to data loss. The proposed method
  can be applied to various applications, such as autonomous driving, robotics,
  and augmented reality, where accurate and robust localization is essential. In
  the future, we will explore the integration of the proposed method with other sensors,
  such as cameras and LiDAR, to further improve the localization accuracy and robustness.
  Additionally, we will investigate the application of the proposed method to other
  datasets and scenarios, such as indoor environments and dynamic scenes, to evaluate
  its generalization capability and robustness.





  \ifCLASSOPTIONcaptionsoff
  \newpage
  \fi

  \bibliographystyle{ieeetr}
  \bibliography{ref}
\end{document}